\documentclass[runningheads]{llncs}
\usepackage{amsmath}
\usepackage[T1]{fontenc}
\usepackage{booktabs}   
\usepackage{multirow}   
\usepackage{array}      
\usepackage{tabularx}   
\usepackage{adjustbox}  
\usepackage{wrapfig}    
\usepackage{graphicx}
\usepackage{placeins}
\usepackage[
  colorlinks=true,
  linkcolor=blue,
  citecolor=blue,
  urlcolor=blue
]{hyperref}

\begin{document}
\title{Robustness of Transformer-Based Fluence Map Prediction Under Clinically Realistic Perturbations}
\titlerunning{Robustness of Transformer-Based Fluence Map Prediction}

\author{
Ujunwa Mgboh\inst{1} \and
Rafi Ibn Sultan\inst{1} \and
Joshua Kim\inst{2} \and
Kundan Thind\inst{2} \and
Dongxiao Zhu\inst{1}
}

\authorrunning{U. Mgboh et al.}

\institute{
Wayne State University, Detroit, MI, USA\\
\email{\{ujunwa.mgboh,rafis,dzhu\}@wayne.edu}
\and
Henry Ford Health, Detroit, MI\\
\email{\{JKIM8,kthind1\}@hfhs.org}
}
\maketitle              
\begin{abstract}
Learning-based fluence map prediction is a promising alternative to iterative inverse planning in intensity-modulated radiation therapy (IMRT), but its robustness under realistic distribution shifts remains unclear. In deployment, models must contend with imaging noise, geometric misalignment, limited supervision, and scanner-dependent domain shifts. We study the robustness of a two-stage transformer-based fluence prediction pipeline, where a first network maps patient anatomy (CT and contours) to a dose distribution and a second network regresses beamlet fluence maps from the predicted dose and beam geometry. Within this unified anatomy $\rightarrow$ dose $\rightarrow$ fluence formulation, we compare fluence-stage encoder backbones with hierarchical, fully global, and hybrid attention, all trained with a physics-informed regression loss that enforces global energy consistency. Robustness is evaluated along four clinically motivated scenarios: geometric setup perturbations, radiometric degradation, reduced training data, and synthetic domain shifts, using a single-institution prostate IMRT dataset with complete CT, contour, dose, and fluence information. We additionally stress-test the anatomy-to-dose stage (CT + contours $\rightarrow$ dose) on two public datasets (OpenKBP-Opt head-and-neck and a CORT prostate case) that provide CT, contours, and dose only. Across stress tests, moderate perturbations yield smooth performance decay, whereas large rotations and severe noise cause sharper drops in both structural similarity and tail-dose stability. Hierarchical transformers such as SwinUNETR exhibit the slowest growth in upper-quartile energy error, indicating improved robustness at similar clean performance and revealing failure modes that are not captured by SSIM alone.

\keywords{
Radiotherapy Planning \and
IMRT Fluence Prediction \and
Transformer Models \and
Robustness Analysis \and
Physics-Informed Learning \and
Clinical Deployment Safety
}
\end{abstract}
\section{Introduction}
\label{sec:introduction}

Clinical radiotherapy must operate under unavoidable uncertainty in imaging, patient setup, and anatomy over time, so fluence prediction models are routinely exposed to distribution shifts at test time. Despite advances in image guidance and quality assurance, residual setup errors, scanner-dependent noise, and protocol-specific variability remain. Conventional planning manages these through geometric margins and robustness constraints~\cite{r1}. When learning-based models are inserted into this pipeline, however, how these uncertainties interact with learned representations becomes less transparent.

Deep learning is increasingly used in radiotherapy planning for dose prediction~\cite{r5} and, more recently, fluence map prediction~\cite{r6,r7}, offering faster planning and reduced inter-operator variability by replacing or accelerating iterative optimization. Yet most evaluations emphasize accuracy on selected test sets and implicitly assume stable data distributions that rarely hold at deployment.

In medical AI more broadly, models trained under idealized conditions are known to degrade under dataset shift, scanner variability, geometric perturbations, and image corruption~\cite{r8,r9}, and large-scale studies show that retrospective accuracy is only weakly correlated with reliability under deployment-time uncertainty~\cite{r12}. These findings motivate robustness auditing and stress testing as prerequisites for responsible clinical adoption~\cite{r13}.

Within radiotherapy, robustness analyses have focused mainly on segmentation and dose prediction, where distribution shifts can cause contour leakage or dose misestimation~\cite{r15,r24}. Direct fluence map prediction introduces a distinct and underexplored risk profile: fluence maps are dense regression targets constrained by beam geometry and energy conservation, so small geometric or radiometric perturbations can induce non-linear changes in modulation that are not obvious from image-similarity metrics yet yield clinically meaningful dosimetric deviations~\cite{r16}. Recent transformer-based fluence predictors such as FluenceFormer~\cite{r17}
combine global or hierarchical attention with physics-informed objectives to
emphasize long-range anatomy. However,
their robustness under clinically realistic uncertainty remains largely
uncharacterized.

In this work, we examine the robustness of deep learning-based fluence map prediction under the main uncertainty sources encountered in IMRT planning: geometric setup errors (patient positioning), radiometric noise and corruption (scanner and acquisition variability), domain shifts across scanners and protocols, and limited annotated training data. Within the FluenceFormer framework~\cite{r17}, we compare several transformer-based fluence predictors along these stress scenarios, quantifying both structural changes in fluence maps and physics-based energy deviations. This analysis reveals how different architectural choices shape degradation curves, delineates operating ranges in which direct fluence prediction remains reliable, and identifies regimes where robustness breaks down in clinically meaningful ways, providing guidance for the safe deployment of learning-based fluence prediction in IMRT planning.

\section{Methods}
\label{sec:methods}

\subsection{Problem formulation and evaluation setup}
\label{subsec:problem}

We model IMRT planning with a two-stage pipeline. Let $I \in \mathrm{R}^{H \times W \times D}$ denote the planning CT volume, where $H$, $W$, and $D$ correspond to the spatial height, width, and depth (number of slices), and $M \in \{0,1\}^{K \times H \times W \times D}$ the $K$-channel anatomical masks defined on the same grid. 

The first stage predicts dose from anatomy,
\begin{equation}
    g_\phi : (I, M) \rightarrow \hat{D},
\end{equation}
and the second stage predicts beamlet fluence maps conditioned on the predicted dose and beam geometry,
\begin{equation}
    f_\theta : (\hat{D}, B) \rightarrow \hat{F},
\end{equation}
where $\hat{D}$ denotes the predicted dose distribution, $B$ encodes beam configuration, and $\hat{F}$ contains per-field fluence maps.

Our robustness analysis covers both $g_\phi$ and $f_\theta$, with emphasis on $f_\theta$ since it directly governs deliverable modulation. For each stage, we quantify performance degradation under geometric, radiometric, domain, and data-scarcity perturbations, using the institutional and public datasets described in Sec.~\ref{subsec:datasets}.

\subsection{FluenceFormer and backbone variants}
\label{subsec:fluenceformer}

In all experiments, $g_\phi$ and $f_\theta$ follow the FluenceFormer framework~\cite{r17}. The fluence stage implements $f_\theta$ as a dense regression model with a transformer encoder and a ReLU output head to enforce non-negativity, using self-attention to capture long-range anatomical context.

We evaluate FluenceFormer backbones SwinUNETR, UNETR, nnFormer, and MedFormer. They share the same two-stage anatomy$\rightarrow$dose$\rightarrow$fluence pipeline, regression head, and supervision, differing only in attention structure (hierarchical windowed, fully global, or hybrid). This isolates how architectural inductive bias affects robustness as perturbation severity increases, rather than emphasizing cross-model ranking.

All transformer variants are trained with the Fluence-Aware Regression (FAR) loss~\cite{r17},
\begin{equation}
\mathcal{L}_{\mathrm{FAR}} =
\alpha \mathcal{L}_{\mathrm{MSE}} +
\beta \mathcal{L}_{\mathrm{Grad}} +
\gamma \mathcal{L}_{\mathrm{Corr}} +
\delta \mathcal{L}_{\mathrm{Energy}},
\end{equation}
where $\mathcal{L}_{\mathrm{MSE}}$ denotes the voxel-wise mean squared error, $\mathcal{L}_{\mathrm{Grad}}$ enforces gradient consistency, $\mathcal{L}_{\mathrm{Corr}}$ measures correlation between predicted and reference fluence, and $\mathcal{L}_{\mathrm{Energy}}$ penalizes deviations from total beam energy. The coefficients $\alpha$, $\beta$, $\gamma$, and $\delta$ control the relative contribution of each term. Together, these components enforce voxel-level accuracy, structural consistency, and physically consistent energy delivery. For context, we also report CNN-based~\cite{r23} and Swin-Direct~\cite{r6} baselines, where Swin-Direct maps CT and contours directly to fluence without explicit beam-geometry encoding; these are included only for reference, not as primary competitors to the FluenceFormer variants.

\subsection{Datasets}
\label{subsec:datasets}

Robustness is evaluated primarily on a single-institution prostate IMRT cohort following the FluenceFormer protocol~\cite{r17}. The dataset comprises 99 anonymized nine-field plans, each with planning CT, anatomical contours, clinical dose, and beam-specific fluence maps exported from Eclipse (Varian), providing full CT/contour/dose/fluence information for the two-stage pipeline.

To assess cross-dataset robustness of the anatomy-to-dose stage $g_\phi$, we additionally stress-test on the OpenKBP head-and-neck dataset~\cite{r19} and the CORT prostate IMRT dataset~\cite{r20}, which provide CT, contours, and dose but no fluence maps. We apply the prostate-trained dose regressor to these cohorts \emph{without} fine-tuning to probe cross-site and cross-organ robustness under a deliberately mismatched setting, restricting evaluation to CT/contour-to-dose prediction on the public datasets and to end-to-end fluence robustness on the institutional cohort.

\subsection{Preprocessing and experimental setup}
\label{subsec:exp_settings}

All datasets (institutional and public) are resampled to a $128 \times 128$ in-plane grid. CT intensities are clipped to a soft-tissue window and normalized to $[0,1]$. Institutional fluence maps with varying native resolutions are upsampled to the same grid and preserved in Monitor Units using a fixed global scaling, enabling physics-aware regression and energy-based evaluation. Dose volumes for anatomy-to-dose regression are likewise resampled and scaled using a consistent global factor.

All models are implemented in PyTorch with MONAI transformer backbones and trained using Adam (learning rate $1 \times 10^{-4}$, batch size $16$) for $50$ epochs per stage.

\subsection{Perturbation and robustness evaluation}

We design four perturbation scenarios to model deployment-time uncertainty rather than adversarial attacks. Translational shifts of $3$--$5$\,mm and in-plane rotations of $2^\circ$--$5^\circ$ approximate residual setup errors~\cite{r21,r22}; additive Gaussian noise ($\sigma=0.05$--$0.20$) and smooth multiplicative bias fields mimic imaging noise and scanner-dependent intensity variations. Data scarcity is simulated by training on reduced fractions of the cohort.

All perturbations are applied at inference on the held-out test set to stress-test both stages of FluenceFormer: anatomy-to-dose prediction $g_\phi$ (public datasets) and dose-conditioned fluence prediction $f_\theta$ (institutional cohort with fluence labels).

\textbf{Geometric robustness.}
Let $X(\mathbf{r})$ denote the input CT (or anatomy channel) at spatial coordinate $\mathbf{r}=(x,y)^\top$. Rigid setup errors are modeled as
\begin{equation}
X_{\text{geom}}(\mathbf{r}) = X\!\left(\mathbf{R}\mathbf{r} + \mathbf{t}\right),
\end{equation}
where $\mathbf{R}$ is a 2D rotation with $\theta\in\{\pm2^\circ,\pm5^\circ\}$ and $\mathbf{t}$ is an in-plane shift with $\|\mathbf{t}\|\in\{3,5\}$\,mm.

\textbf{Radiometric robustness.}
Scanner noise is modeled as
\begin{equation}
X_{\text{noise}}(\mathbf{r}) = X(\mathbf{r}) + \epsilon(\mathbf{r}), \quad
\epsilon(\mathbf{r}) \sim \mathcal{N}(0,\sigma^2),
\end{equation}
with $\sigma\in\{0.05,0.10,0.15,0.20\}$.

\textbf{Data efficiency.}
Data scarcity is simulated by retraining on subsets $\mathcal{D}_\alpha\subset\mathcal{D}$ with $\alpha\in\{0.25,0.50,0.75\}$.

\textbf{Domain robustness.}
Cross-scanner variability is approximated by
\begin{equation}
X_{\text{domain}}(\mathbf{r}) = b(\mathbf{r})\,X(\mathbf{r}) + c,
\end{equation}
where $b(\mathbf{r})$ is a smooth bias field and $c$ a global offset.

Performance is evaluated using Structural Similarity Index Measure (SSIM) and energy error. To capture clinically relevant failure cases, energy error is summarized using its 75th percentile $E_{\mathrm{err}}^{Q_{75}}$ across patients.


\section{Results}
\label{sec:results}

All trends were consistent across patients, and within-model changes under increasing perturbation were statistically significant (paired Wilcoxon signed-rank test, $p<0.05$). We focus on robustness degradation within each model as stress increases rather than cross-model ranking.
\subsection{Geometric robustness to setup errors}

\begin{figure}[!htbp]
\centering
\includegraphics[width=\linewidth]{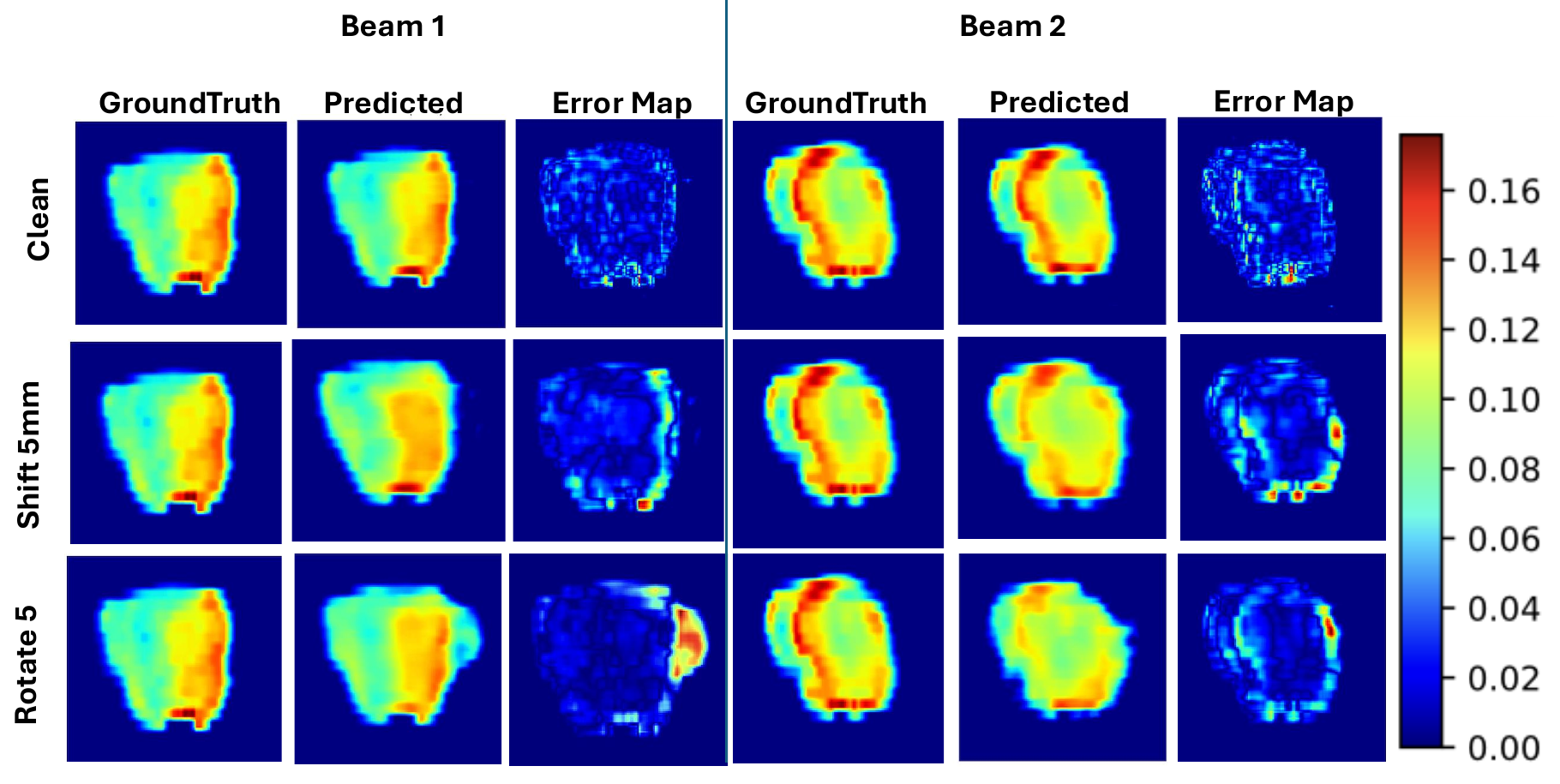}
\caption{\textbf{Qualitative geometric robustness.}
Ground-truth fluence maps (repeated for visual reference) remain invariant, as geometric perturbations are applied only at inference.}
\label{fig:geo_qualitative}
\end{figure}

Table~\ref{tab:geo_all} and Fig.~\ref{fig:geo_qualitative} summarize robustness to translational and rotational misalignment using SSIM and $E_{\mathrm{err}}^{Q_{75}}$. Under shifts from $\pm3$\,mm to $\pm5$\,mm, transformer backbones remain stable (e.g., SwinUNETR: SSIM $\approx 0.75 \rightarrow 0.74$, $E_{\mathrm{err}}^{Q_{75}} \approx 5\%$), while the CNN baseline shows similar SSIM but noticeably higher tail energy error. Rotations are more damaging for all models: SwinUNETR shows only moderate growth in $E_{\mathrm{err}}^{Q_{75}}$ from $2^\circ$ to $5^\circ$, whereas CNN attains the largest tail errors. Overall degradation is smooth rather than catastrophic, with hierarchical attention (SwinUNETR) consistently yielding the slowest increase in $E_{\mathrm{err}}^{Q_{75}}$ compared to CNN and fully global variants.

\begin{table}[!htbp]
\centering
\scriptsize
\setlength{\tabcolsep}{2pt}
\caption{\textbf{Geometric robustness to setup errors.}
SSIM and upper-quartile energy error $E_{\mathrm{err}}^{Q_{75}}$ under translational (3, 5\,mm) and rotational (2$^\circ$, 5$^\circ$) perturbations. 
The upper block (\emph{SwinUNETR, UNETR, nnFormer, MedFormer}) shows FluenceFormer-based backbones sharing the same two-stage anatomy$\rightarrow$dose$\rightarrow$fluence pipeline and Fluence-Aware Regression loss; the lower block (\emph{CNN, Swin\_Direct}) lists prior single-stage fluence baselines. 
Higher SSIM and lower $E_{\mathrm{err}}^{Q_{75}}$ are better; \textbf{bold} entries mark, \emph{within each model and perturbation family}, the level with smaller performance degradation rather than the best model overall.}
\label{tab:geo_all}
\resizebox{0.9\textwidth}{!}{%
\begin{tabular}{lcccccccc}
\toprule
& \multicolumn{4}{c}{\textbf{Shift}} & \multicolumn{4}{c}{\textbf{Rotation}} \\
\cmidrule(lr){2-5}\cmidrule(lr){6-9}
\textbf{Model} &
\multicolumn{2}{c}{3 mm} &
\multicolumn{2}{c}{5 mm} &
\multicolumn{2}{c}{2$^\circ$} &
\multicolumn{2}{c}{5$^\circ$} \\
\cmidrule(lr){2-3}\cmidrule(lr){4-5}\cmidrule(lr){6-7}\cmidrule(lr){8-9}
& SSIM$\uparrow$ & $E_{\mathrm{err}}^{Q_{75}}\downarrow$ &
  SSIM$\uparrow$ & $E_{\mathrm{err}}^{Q_{75}}\downarrow$ &
  SSIM$\uparrow$ & $E_{\mathrm{err}}^{Q_{75}}\downarrow$ &
  SSIM$\uparrow$ & $E_{\mathrm{err}}^{Q_{75}}\downarrow$ \\
\midrule
\multicolumn{9}{l}{\textit{FluenceFormer-based backbones}} \\
SwinUNETR        
& \textbf{0.75$\pm$0.04} & \textbf{5.67}
& 0.74$\pm$0.04 & 5.03
& \textbf{0.75$\pm$0.07} & \textbf{7.32}
& 0.72$\pm$0.05 & 9.33 \\
UNETR            
& \textbf{0.65$\pm$0.04} & \textbf{6.26}
& 0.63$\pm$0.04 & 5.81
& \textbf{0.66$\pm$0.05} & \textbf{9.25}
& 0.62$\pm$0.04 & 11.51 \\
nnFormer         
& \textbf{0.56$\pm$0.05} & \textbf{7.43}
& 0.54$\pm$0.04 & 8.21
& \textbf{0.57$\pm$0.04} & \textbf{10.37}
& 0.53$\pm$0.04 & 10.64 \\
MedFormer        
& \textbf{0.63$\pm$0.03} & \textbf{7.32}
& 0.61$\pm$0.04 & 7.95
& \textbf{0.64$\pm$0.04} & \textbf{11.47}
& 0.60$\pm$0.03 & 11.80 \\
\midrule
\multicolumn{9}{l}{\textit{Baseline fluence predictors}} \\
CNN              
& 0.62$\pm$0.05     & \textbf{10.45}
& \textbf{0.63$\pm$0.05} & 14.45
& \textbf{0.64$\pm$0.08} & \textbf{17.70}
& 0.64$\pm$0.06 & 20.86 \\
Swin\_Direct      
& \textbf{0.62$\pm$0.04} & \textbf{9.21}
& 0.61$\pm$0.04 & 11.65
& \textbf{0.64$\pm$0.06} & \textbf{14.86}
& 0.64$\pm$0.06 & 18.52 \\
\bottomrule
\end{tabular}%
}
\end{table}

\subsection{Robustness to image quality degradation}

\begin{figure}[!htbp]
  \centering
  \includegraphics[width=1.0\linewidth]{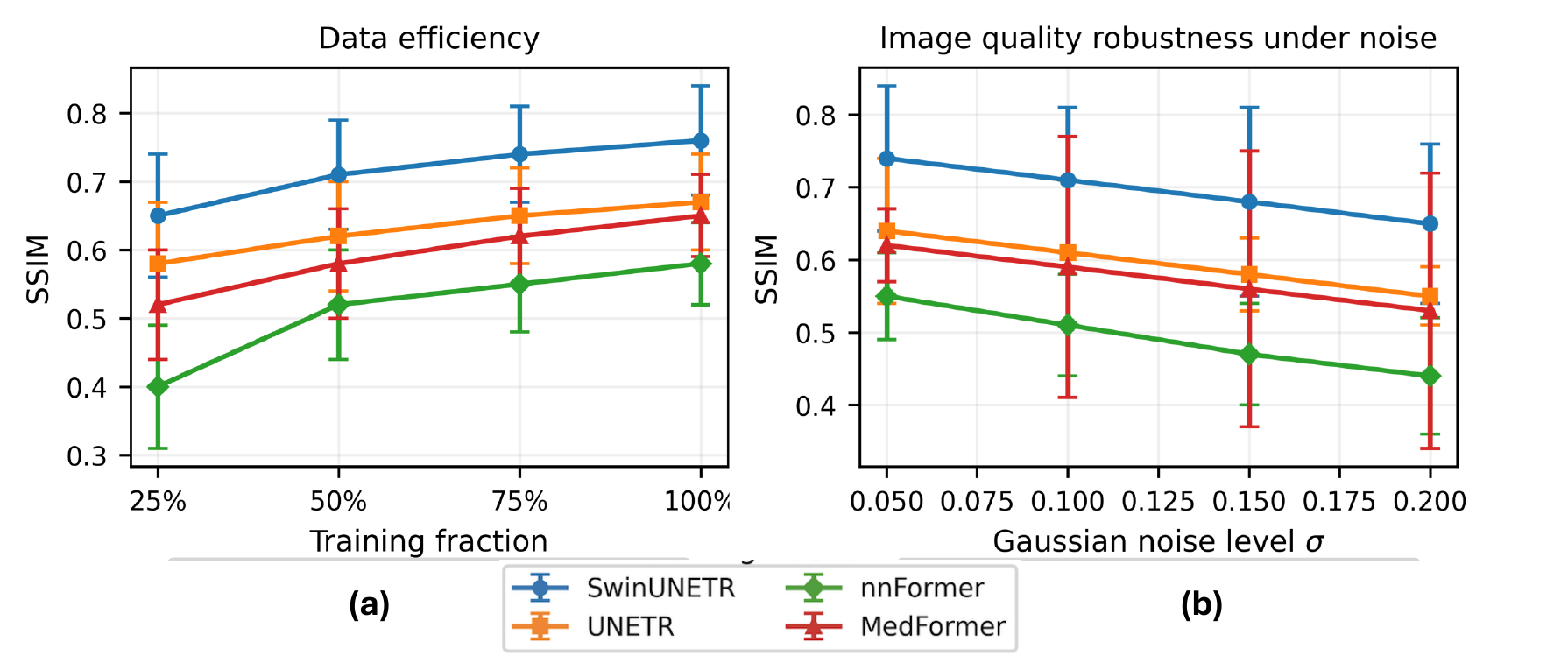}
  \caption{\textbf{Data efficiency and radiometric robustness.}
  (a) SSIM versus training fraction for FluenceFormer backbones, showing monotonic gains with more supervision.
  (b) SSIM versus Gaussian noise level $\sigma$ on CT input, showing smooth degradation under increasing noise.
  Error bars denote standard deviation across test cases.}
  \label{fig:data_noise_robustness}
\end{figure}

Radiometric noise perturbs voxel intensities without altering anatomy, making it a clean probe of intensity sensitivity. Table~\ref{tab:noise_robustness} and Fig.~\ref{fig:data_noise_robustness}(b) show smooth, monotonic SSIM decay as Gaussian noise increases from $\sigma=0.05$ to $0.20$ for all backbones. SwinUNETR, UNETR, and MedFormer lose $\sim$10--15\% SSIM over this range, whereas nnFormer is most sensitive, with a drop of over 20\%, indicating reduced robustness to radiometric corruption.

\begin{table}[!htbp]
\centering
\scriptsize
\setlength{\tabcolsep}{2pt}
\caption{\textbf{Image quality robustness (FluenceFormer backbones).}
SSIM under Gaussian noise with standard deviation $\sigma$.}
\label{tab:noise_robustness}
\resizebox{0.75\textwidth}{!}{%
\begin{tabular}{lcccc}
\toprule
\textbf{Backbone} & $\sigma=0.05$ & $\sigma=0.10$ & $\sigma=0.15$ & $\sigma=0.20$ \\
\midrule
SwinUNETR & \textbf{0.74$\pm$0.10} & 0.71$\pm$0.10 & 0.68$\pm$0.13 & 0.65$\pm$0.11 \\
UNETR     & \textbf{0.64$\pm$0.10} & 0.61$\pm$0.10 & 0.58$\pm$0.05 & 0.55$\pm$0.04 \\
nnFormer  & \textbf{0.55$\pm$0.06} & 0.51$\pm$0.07 & 0.47$\pm$0.07 & 0.44$\pm$0.08 \\
MedFormer & \textbf{0.62$\pm$0.05} & 0.59$\pm$0.18 & 0.56$\pm$0.19 & 0.53$\pm$0.19 \\
\bottomrule
\end{tabular}%
}
\end{table}

\vspace{-2mm}
\subsection{Data efficiency}
\label{subsec:dataeff}

All FluenceFormer variants improve monotonically with more training data (Table~\ref{tab:data_efficiency}, Fig.~\ref{fig:data_noise_robustness}(a)). SwinUNETR is the most data-efficient, achieving the highest SSIM and lowest $E_{\mathrm{err}}^{Q_{75}}$ even at 25\% of the data, while nnFormer is the most data-hungry, requiring full supervision to approach comparable SSIM and still retaining higher tail energy error. Overall, the smooth scaling curves indicate graceful degradation under data scarcity, with hierarchical attention offering the best robustness data trade-off.

\begin{table}[!htbp]
\centering
\scriptsize
\setlength{\tabcolsep}{2pt}
\caption{\textbf{Data efficiency and scaling.}
SSIM and upper-quartile energy error $E_{\mathrm{err}}^{Q_{75}}$ as a function of training data fraction.}
\label{tab:data_efficiency}
\resizebox{0.8\textwidth}{!}{%
\begin{tabular}{llcccc}
\toprule
\textbf{Backbone} & \textbf{Metric} & \textbf{25\%} & \textbf{50\%} & \textbf{75\%} & \textbf{100\%} \\
\midrule
\multirow{2}{*}{SwinUNETR}
& SSIM$\uparrow$ &
0.65$\pm$0.09 & 0.71$\pm$0.08 & 0.74$\pm$0.07 & \textbf{0.76$\pm$0.08} \\
& $E_{\mathrm{err}}^{Q_{75}}\downarrow$ &
10.56 & 7.32 & 5.43 & \textbf{4.67} \\
\midrule
\multirow{2}{*}{UNETR}
& SSIM$\uparrow$ &
0.58$\pm$0.09 & 0.62$\pm$0.08 & 0.65$\pm$0.07 & \textbf{0.67$\pm$0.07} \\
& $E_{\mathrm{err}}^{Q_{75}}\downarrow$ &
12.56 & 10.89 & 10.56 & \textbf{8.67} \\
\midrule
\multirow{2}{*}{nnFormer}
& SSIM$\uparrow$ &
0.40$\pm$0.09 & 0.52$\pm$0.08 & 0.55$\pm$0.07 & \textbf{0.58$\pm$0.06} \\
& $E_{\mathrm{err}}^{Q_{75}}\downarrow$ &
12.78 & 11.89 & 11.56 & \textbf{10.34} \\
\midrule
\multirow{2}{*}{MedFormer}
& SSIM$\uparrow$ &
0.52$\pm$0.08 & 0.58$\pm$0.08 & 0.62$\pm$0.07 & \textbf{0.65$\pm$0.06} \\
& $E_{\mathrm{err}}^{Q_{75}}\downarrow$ &
12.26 & 11.92 & 11.45 & \textbf{10.67} \\
\bottomrule
\end{tabular}%
}
\end{table}

\begin{figure}[!htbp]
    \centering
    \resizebox{1.06\textwidth}{!}{%
        \includegraphics{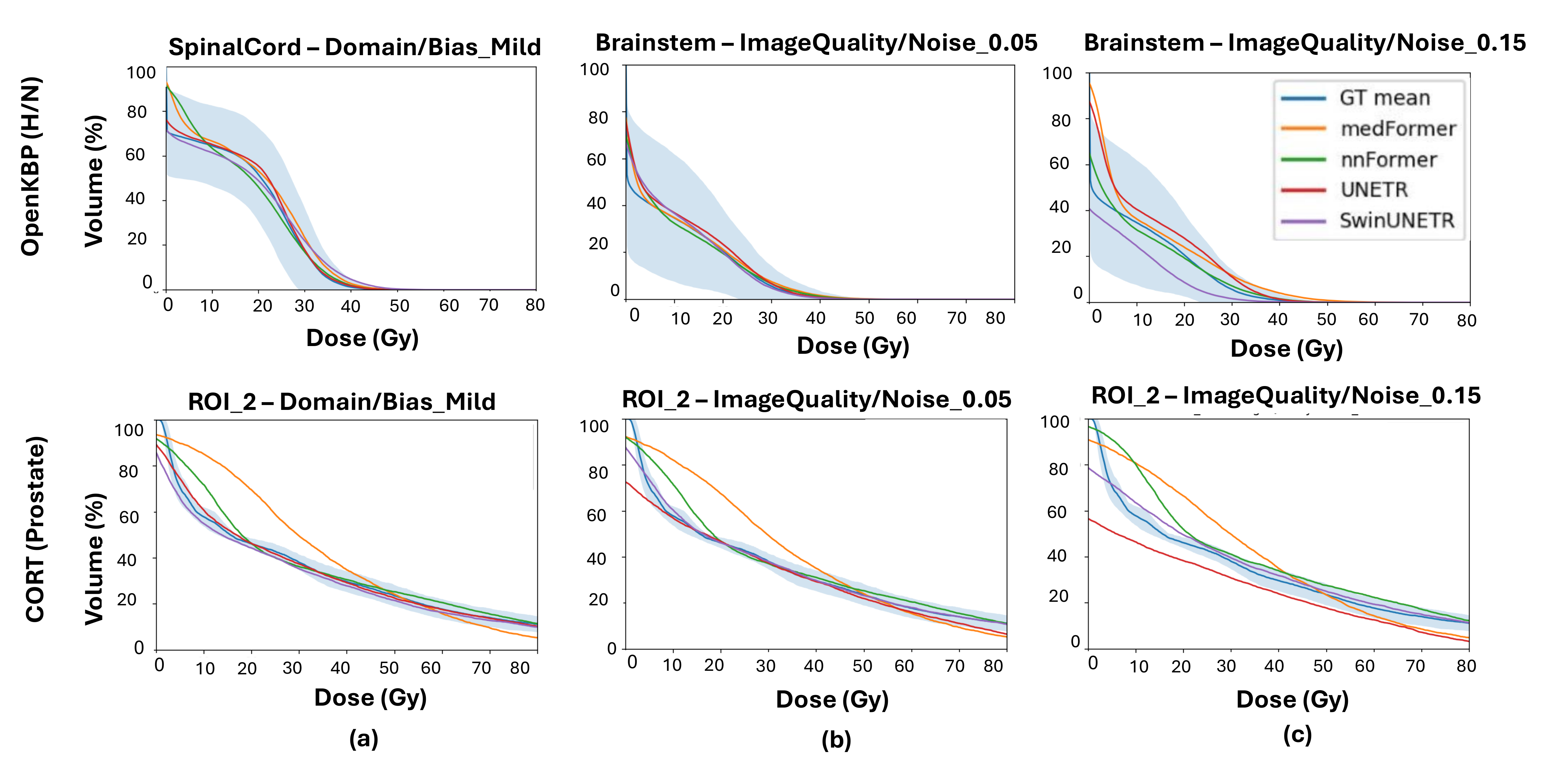}%
    }
    \caption{\textbf{DVH robustness on public datasets.}
    DVH curves for a representative organ-at-risk on the OpenKBP head-and-neck cohort (top row) and a matched prostate ROI on the CORT dataset (bottom row) under (a) mild bias-field perturbation, (b) low Gaussian noise ($\sigma=0.05$), and (c) higher noise ($\sigma=0.15$). Shaded regions denote ground-truth mean$\pm$std; colored curves show different backbones.}
    \label{fig:public_dvh}
\end{figure}


\subsection{Domain robustness}

Bias-field perturbations primarily affect dosimetric stability rather than structural fidelity. As shown in Table~\ref{tab:domain_robustness}, SSIM remains similar across mild and severe bias, but $E_{\mathrm{err}}^{Q_{75}}$ increases with bias strength. SwinUNETR consistently yields the lowest tail energy error under domain shift, CNN and SwinUNETR-Direct  maintain high SSIM yet exhibit large tail errors, indicating visually plausible but dosimetrically unreliable predictions.

\begin{table}[!htbp]
\centering
\scriptsize
\setlength{\tabcolsep}{1.5pt} 
\caption{\textbf{Domain robustness under bias-field artifacts.}
SSIM and upper-quartile energy error $E_{\mathrm{err}}^{Q_{75}}$ under mild
and severe bias fields.}
\label{tab:domain_robustness}
\resizebox{0.85\textwidth}{!}{%
\begin{tabular}{lcccc}
\toprule
\multirow{2}{*}{\textbf{Backbone}} &
\multicolumn{2}{c}{\textbf{Bias mild}} &
\multicolumn{2}{c}{\textbf{Bias severe}} \\
& SSIM$\uparrow$ & $E_{\mathrm{err}}^{Q_{75}}\downarrow$ &
  SSIM$\uparrow$ & $E_{\mathrm{err}}^{Q_{75}}\downarrow$ \\
\midrule
SwinUNETR        & \textbf{0.75$\pm$0.12} & \textbf{7.33} & 0.73$\pm$0.12 &  9.65 \\
UNETR            & \textbf{0.66$\pm$0.11} & 10.17        & 0.64$\pm$0.10 & \textbf{10.04} \\
nnFormer         & \textbf{0.56$\pm$0.05} & \textbf{9.65} & 0.52$\pm$0.13 & 11.89 \\
MedFormer        & \textbf{0.63$\pm$0.04} & 10.70        & 0.60$\pm$0.03 & \textbf{9.65} \\
\midrule
CNN              & \textbf{0.66$\pm$0.13} & 15.44        & \textbf{0.66$\pm$0.13} & \textbf{14.94} \\
SwinUNETR\_Direct& \textbf{0.66$\pm$0.12} & \textbf{15.73} & \textbf{0.66$\pm$0.11} & 16.97 \\
\bottomrule
\end{tabular}%
}
\end{table}
\subsection{External robustness of dose prediction on public datasets}

We assess cross-dataset robustness of the dose-prediction stage by applying the prostate-trained regressor to the OpenKBP head-and-neck cohort and the CORT prostate dataset, recomputing DVHs under the same stress scenarios. As shown in Fig.~\ref{fig:public_dvh}, spinal cord and brainstem DVHs on OpenKBP (top) and the prostate ROI on CORT (bottom) degrade smoothly from mild bias (a) through low (b) to higher (c) noise. Across both datasets, SwinUNETR and UNETR stay closest to the ground-truth DVH band, while nnFormer and MedFormer deviate more under stronger perturbations. This indicates that the anatomy-to-dose mapping generalizes under moderate domain shift, even though end-to-end fluence robustness cannot be assessed on datasets without fluence maps.

\section{Discussion and Conclusion}
\label{sec:discussion}

We conducted a systematic stress test of transformer-based fluence prediction, emphasizing robustness under clinically motivated uncertainty rather than clean-condition accuracy. Across geometric, radiometric, data-scarcity, and domain-shift scenarios, robustness emerges from the interplay between architectural inductive bias and physics-informed supervision.

Geometric experiments confirm that rotations are more disruptive than translations (Table~\ref{tab:geo_all}), consistent with beam--anatomy physics. Under rotational setup errors, fully global attention (e.g., UNETR) shows sharper drops in SSIM and larger increases in upper-quartile energy error, whereas SwinUNETR degrades more smoothly. This suggests that hierarchical windowed attention improves resilience to misalignment by limiting the propagation of spatial errors.

Radiometric noise produces smooth SSIM decay across all backbones (Table~\ref{tab:noise_robustness}), but sensitivity varies: nnFormer degrades most, while SwinUNETR and MedFormer are more tolerant, consistent with multi-scale representations that suppress high-frequency noise.

Domain-shift experiments highlight a gap between visual and physical robustness. Under bias fields (Table~\ref{tab:domain_robustness}), CNN and SwinUNETR-Direct retain high SSIM yet incur large tail energy errors, yielding visually plausible but dosimetrically unsafe predictions. In contrast, FluenceFormer variants trained with the fluence-aware regression loss better control energy drift, with SwinUNETR showing the slowest growth in $E_{\mathrm{err}}^{Q_{75}}$, underscoring the value of physics-informed supervision.

Data-efficiency results (Table~\ref{tab:data_efficiency}) show that while all models benefit from more data, they scale differently: nnFormer is strongly data-dependent, whereas SwinUNETR maintains higher SSIM and lower tail energy error even at 25\% of the data, indicating improved robustness in limited-data regimes.

External experiments on OpenKBP and CORT (Fig.~\ref{fig:public_dvh}) show that the anatomy-to-dose stage generalizes under cross-dataset shift, with DVHs degrading smoothly and remaining clinically acceptable under moderate perturbations. 

These findings emphasize that robustness evaluation, rather than architectural novelty alone, is critical for assessing clinical reliability. SSIM alone is insufficient, as it may mask physically inconsistent predictions, whereas energy-based metrics better capture clinically relevant failure modes.

Overall, transformer-based fluence prediction can be robust within realistic IMRT operating regimes, but systematic stress testing and physics-informed objectives are essential to ensure safe and reliable deployment.
%
%

%
%
%
%

\end{document}